\UseRawInputEncoding

\documentclass[journal]{IEEEtran}
\usepackage{cite}

\usepackage{graphicx}
\ifCLASSINFOpdf
\else
\fi
\ifCLASSOPTIONcompsoc
  \usepackage[caption=false,font=normalsize,labelfont=sf,textfont=sf]{subfig}
\else
  \usepackage[caption=false,font=footnotesize]{subfig}
\fi
\usepackage{multirow}

\begin{document}
%
\title{A Logical Neural Network Structure With More Direct Mapping From Logical Relations}
%
%
%

\author{Gang~Wang
\thanks{G. Wang is with the School of Computer Science and Technology, Taiyuan University of Technology, Taiyuan, China
(e-mail: f\_lag@buaa.edu.cn.)}
}

%
%

\markboth{}
{Wang \MakeLowercase{\textit{et al.}}: A Logical Neural Network Structure with more direct mapping from Logical Relations}
%



\maketitle

\begin{abstract}
Logical relations widely exist in human activities. Human use them for making judgement and decision according to various conditions, which are embodied in the form of \emph{if-then} rules. As an important kind of cognitive intelligence, it is prerequisite of representing and storing logical relations rightly into computer systems so as to make automatic judgement and decision, especially for high-risk domains like medical diagnosis. However, current numeric ANN (Artificial Neural Network) models are good at perceptual intelligence such as image recognition while they are not good at cognitive intelligence such as logical representation, blocking the further application of ANN. To solve it, researchers have tried to design logical ANN models to represent and store logical relations. Although there are some advances in this research area, recent works still have disadvantages because the structures of these logical ANN models still don't map more directly with logical relations which will cause the corresponding logical relations cannot be read out from their network structures. Therefore, in order to represent logical relations more clearly by the neural network structure and to read out logical relations from it, this paper proposes a novel logical ANN model by designing the new logical neurons and links in demand of logical representation. Compared with the recent works on logical ANN models, this logical ANN model has more clear corresponding with logical relations using the more direct mapping method herein, thus logical relations can be read out following the connection patterns of the network structure. Additionally, less neurons are used.
\end{abstract}


%
\IEEEpeerreviewmaketitle

\section{Introduction}
\label{Introduction}
%
%
%
%
\IEEEPARstart{A}{NN} (Artificial Neural Network) model, as one of major AI techniques, has been used in many aspects of people's living and working, and this technique brings great influence on the society increasingly. It has been successfully used in perceptual intelligence such as image classification \cite{Sun, Parkhi, He, Kermany}, speech recognition \cite{Mohamed, Gehring}. However, these current ANN models, which are essentially function approximators, are not good at cognitive intelligence like logical representation and reasoning \cite{wired, Garcez1, LeCun, Francois, Garcez2, Garnelo}. Cognitive intelligence is another important aspect of intelligence besides perceptual intelligence. As an important ability, higher cognitive intelligence makes human more intelligent and better than other animals. The application range of these current ANN models in practice also verifies the above opinion. For example, in the domain of advanced medical diagnosis, most of the application of ANN appears in medical image recognition like breast cancer \cite{Kermany, Rajaraman, Miotto}, but there are few impressive advances in the research and industrial application about using ANN to make diagnosis with medical knowledge which contains various causal relations between symptoms and diseases \cite{Miotto}. With regard to this medical knowledge, the knowledge is expressed in the form of the set of \emph{if-then} rules between symptoms and diseases, and the doctor determines which diseases the patient catches by processing the judgment according to various symptoms in his brain. On the contrast, although ANN models are developed by mimicking the biological brain so as to reach human intelligence, these current ANN models still fulfil partial intelligent functions of the biological brain. It means these current ANN models are good at image classification and speech recognition as kinds of perceptual intelligence while they are not good at logical representation and reasoning as a kind of cognitive intelligence. Rich logical representation and reasoning is an important ability for human and an indispensable function for the brain. If we want the intelligent ability of ANN models moves towards the high intelligence ability of biological brain like the doctors, it is necessary to fulfil the function of the logical representation and reasoning within the artificial neural network structure. These current ANN models work as function approximators, and thus make these models not good at logical representation due to its undirect structural mapping which try to use numeric approximation to express symbolic logic who is the other different mathematical branch. Besides, current numeric ANN models lack interpretability or explainability of its network structure. What do the neurons and links in the neural network represent when an numeric ANN models is used in a domain like medical diagnosis? It is important to know what knowledge an ANN model represent, especially in high-risk domains like medical diagnosis referring to life safety \cite{Miotto, Csiszar}.

With regard to the interpretable techniques of machine learning model, they can generally be divided into two different ways \cite{Mengnan}: designing interpretable models and post-hoc interpretation, i.e. interpreting what domain knowledge its structure represents for a learned model. Research on the interpretation of ANN models from the two ways has been carried out. By comparison with two ways specially in respect of this brain-inspired computing, it is more natural of designing interpretable ANN models since there is no independent interpreting organ in the biological brain. Furthermore, interpreting ANN model is not a thorough solution since this way needs providing an outer interpreter additionally, and it doesn't solve the interpretation of ANN model itself yet. Saying in Chinese old words, the way of interpreting ANN model is a solution to "address the symptom rather than the cause". Therefore, in order to avoid the black-box characteristic of ANN models and to fulfill more functions of biological brain including cognitive intelligence, an interpretable ANN model is required. It will extend the application range of ANN models like medical diagnose aforementioned, and it can bring great economical and social value.

\begin{figure}[!ht]
\begin{center}
\includegraphics[width=0.9\columnwidth]{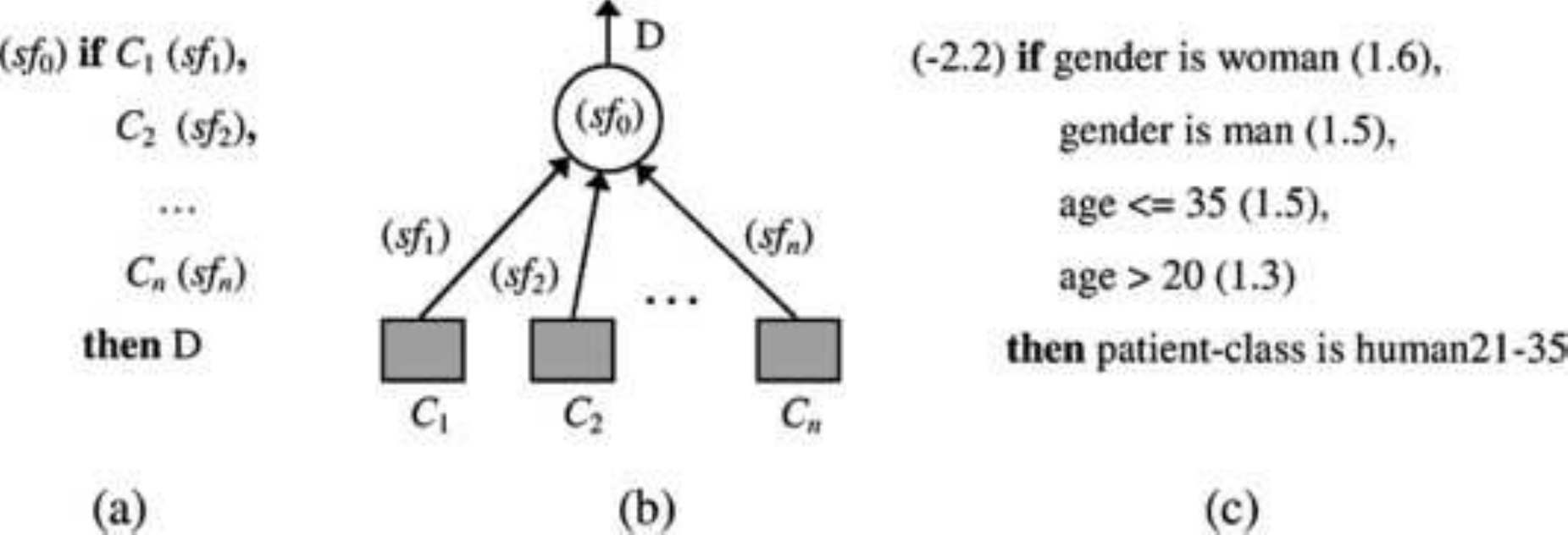}
\caption{Neurule: a kind of LNN structure to represent logical relations. (a) form of a neurule, (b) a neurule as an adaline unit, (c) an example neurule.}
\label{neurule}
\end{center}
\end{figure}

In order to make ANN good at cognitive intelligence and have the ability of deal with the knowledge like the medical knowledge aforementioned, researchers have tried to design new neural network models for addressing logical relations, which are called LNN (Logical Neural Network) models. On the other hand, current ANN models whose internal working principles are function approximators are called NNN (Numeric Neural Network) models. In every day of living and working, human need unstoppably make judgement and decision according to various specific conditions, which are generally embodied in the form of \emph{if-then} logical relations. Human's brains memory various logical relations to make human address more and more complex things like medical diagnosis. Therefore, addressing logical relations is one important part of cognitive intelligence and thus also an indispensable brain function. How to design ANN models to have the ability of addressing logical relations is the unavoidable challenging and valuable work for academic and industrial researchers with the aim of fulfilling the intelligent functions of biological brains and thus achieving human intelligence. To design a LNN model, the first thing to do is how to represent logical relations by using neural network structure or store them in the format of neural network structure. More specifically, how are logical relations represented by using the only two kinds of the components in the neural network model (neurons and links) and constructing the neural network structure? How are the neural network structure designed including the logical meanings of neurons and links and the connection patterns of links betweens neurons to store the information of the represented logical relations? There have been several methods and applications proposed \cite{Besold, Mandziuk, Towell, Valiant, Garcez, Penning, Bowman, Gallant, Wang, Wang1, Ioannis, Csiszar}. Early works of designing logical neural network models are presented in \cite{Gallant, Towell, Garcez, Penning}. These models, named KBNN (Knowledge-Based Neural Network) or Connectionist Expert Systems, firstly define AND/OR neurons and positive and negative links according to the basic logical connectives of the proposition logic type. Then these models represent complex logical relations by the composition of these defined logical neurons and logical links. According to logical relations in a rule library in an expert system, a logical neural network is constructed using these basic AND/OR neurons and links. The logical neural network is interpretable itself. From the inner network structure, it can been what logical relations the neural network represents in the view of the basic component level of the neurons and the links. The type of KBNN can map directly from logical relations into the neural network structure. It represents and stores logical relations by the form of the connection patterns of the neural network structure. Another LNN model named Neurule is presented in \cite{Ioannis, Jim} relatively recently. In the model,  a component neurule is designed to represent a logical relation as shown in Figure~\ref{neurule}. The conditions of a logical relation in the form of \emph{if-then} rules map the inputs of the neurule, and the conclusion maps the output of the neurule. The input can take a value from the following value set: [True, False, Unknown]. The weights of the unit are called the significance factors to represent the effects of the conditions on the conclusion. Then a neurule base consists of a number of neurules to represent the library of logical relations. However, the structure of this ANN model still doesn't represent logical relations more directly, in which logical relations stored in the network structure can not been easily and intuitively recognized. The quite recent work \cite{Csiszar} presents a LNN model by using continuous logic to design logical neurons in the hidden layers with the aim of trying to improve the interpretability of ANN model. Although this model tries to integrate the logic theory into the technique of neural network so as to improve the interpretability of ANN structures, it is still based on the fixed structure of FNN (Forward Neural Network), which can not be flexible to represent a number of various logical relations. Different sets of logical relations will map different network structures. The fixed structure of this model will affect the interpretability of the ANN model on logic representation since the work try to map different sets of logical relations into a single fixed neural network structure.

In order to make the network structure of ANN more directly mapped from logical relations and thus interpret logical relations stored in the network structure by self, this paper presents a logic-interpretable neural network model. This work tries to intemperate logic into neural network model by the way of mapping directly logical relations of propositional type into neural network structure. This logical ANN model defines new neurons and links between neurons to represent basic logical relations. Then the neural network can be dynamically constructed by adding neurons and links according to the set of logical relations rather than the fixed network structure with the unvaried connection patterns. Compared with the two recent ANN models aforementioned, this logical neural network model can interpret by itself, i.e., that it can be directly seen what logical relations are represented by the neural network structure from the neurons and links. In other words, the logical relations can be read out from the neural network structure reversely after this LNN model are constructed according to them. Due to the its self-interpretability of this ANN model on logical representation, we can extend ANN into the domains in which the interpretability is crucial such as medical diagnose aforementioned. These domains demand the clearness of ANN models on logical representation, users need to know what logical relations are exactly rather than ambiguously represented by the neural network structures including neurons and links before making decision using ANN techniques. Furthermore, this LNN model can be further developed with the addition of other features including the probabilistic feature to deal with
uncertainty and the dynamical feature for automatic network construction. This paper only discusses the logical feature of the model in details and makes comparison with recent advances on LNN models. The other features of the model can be seen in the works \cite{Wang, Wang1} if interested.

This paper is organized as follows: Section~\ref{components of LNN} describes the logical neural network presented in the paper. The neuron and link in this logical ANN model are correspondingly defined so as to represent basic logical relations of propositional type. Based on these logical neural components, the neural network can be correspondingly constructed according to logical relations. Due to the direct mapping from logical relations to neural network structure, this LNN model is self-interpretable on logical representation, in whose network structure we can see what logical relations are represented. Section~\ref{structure} describes the neural network structure of logical relations constructed based on these novel logical neurons and links. the neural network of this LNN model will be constructed according to the examples of logical relations used in constructing the neurule base\cite{Ioannis}. Through the comparison of these two LNN models in the case of representing the same logical relations, the result will show that it is more easy and intuitive of interpreting logical relations from the neural network structure of the presented LNN model. Section~\ref{Experiment} will carry out the experiments to test whether this LNN model can represent logical relations. Different datasets from various domains are used to show the practicability and university of this LNN model. Then an demonstration of using this LNN model is given to present the advantages of this LNN model in logical representation due to direct mapping. By graphically depicting the neural network structure inside the model, Section~\ref{Experiment} will intuitively show this direct mapping characteristic from logical relations to the neural network structure. Finally, Section~\ref{Conclusion} concludes.
\section{Design of the components of logical neural network: neurons and links}
\label{components of LNN}
\begin{figure}[!ht]
\begin{center}
\includegraphics[width=0.9\columnwidth]{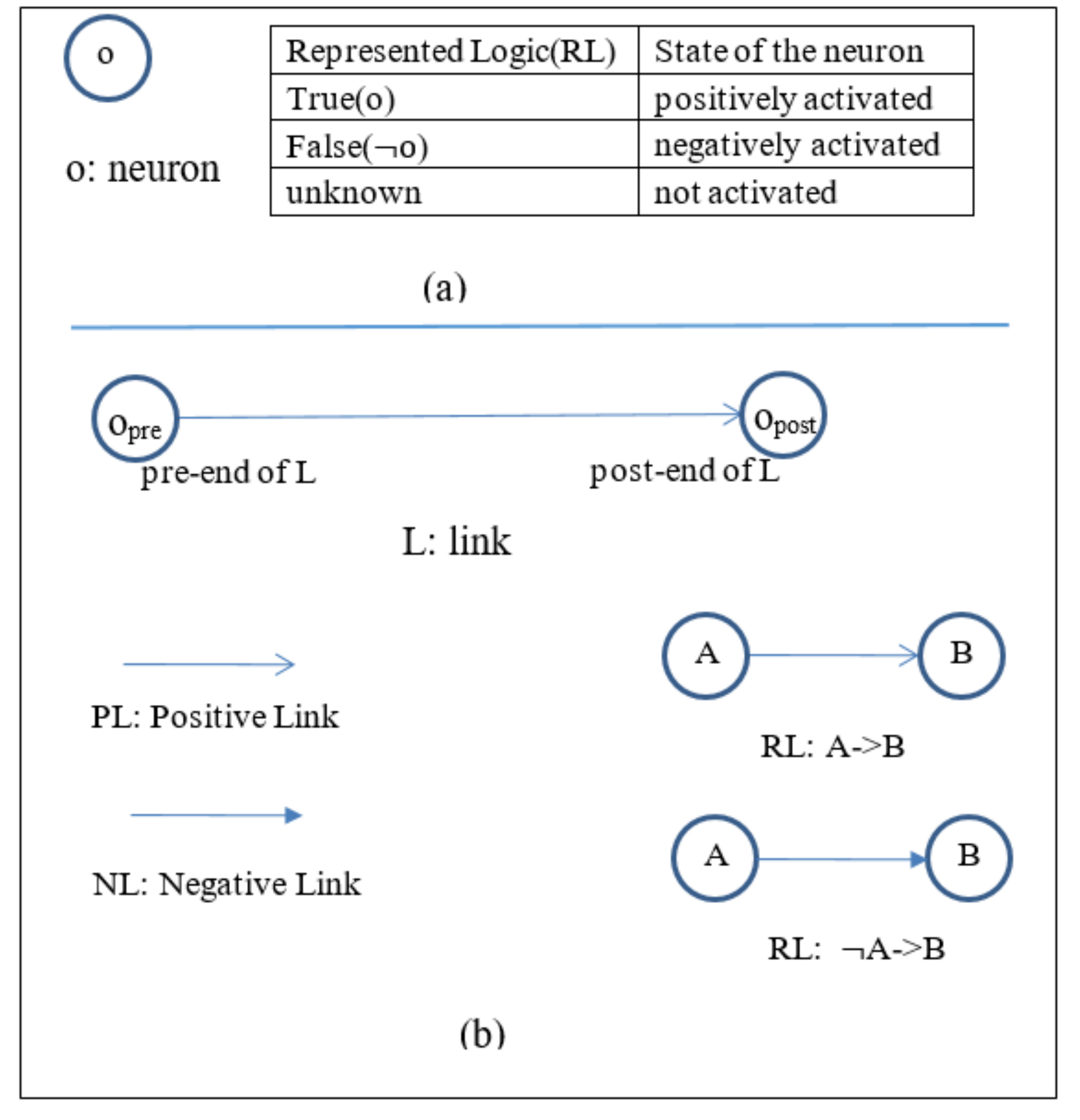}
\caption{Using two components of  ANN for logical representation}
\label{ANN components}
\end{center}
\end{figure}

Aiming to make the neural network structure of the LNN model map directly to logical relations, the paper designs the model from the basic bottom class, i.e. it redefines the basic components: neurons and links according to the requirement to represent logical relations, which are different from those components in numeric neural network model so as to make self-interpretability of the neural network structure for logical relations. First, the neuron o is defined to take two values to correspond True and False in Boolean Logic. The neuron in this LNN model is used for representing a thing which is an object or an event, and is defined by users. When the thing happens and is perceived by the neural network model, the neuron will be positively activated and take the value of representing True. In opposition, when the thing doesn't happens and is perceived by the neural network model, the neuron will be negatively activated and take the other value of representing False. When the neuron is not activated, the paper sets this state of the neuron to represent that the state of the represented thing is unknown. The definition of the neuron in this LNN model for logical representation is shown Figure~\ref{ANN components}a.

After using a neuron to represent a thing and defining the state of the neuron to represent the state of the thing, the next work is design how to represent the logical relations between things by using neural network. It is a natural way of using neurons to represent things while using links between neurons to represent logical relations between things, thus the link L between neurons is used to represent the logical relation $\rightarrow$ between the things. As shown in Figure~\ref{ANN components}b, the left neuron connected by L is called the pre-end neuron while the right neuron connected by L is called the post-end neuron. When the pre-end neuron is in the activated state, the link will be activated. In order to represent logical relations, the links of the model are divides two types. The Positive Link (PL) is used to represent the logical relation $\rightarrow$ while the Negative Link (NL) is used to represent the logical relation $\neg\rightarrow$ such as $\neg A\rightarrow B$.

\begin{figure}[!ht]
\begin{center}
\includegraphics[width=0.9\columnwidth]{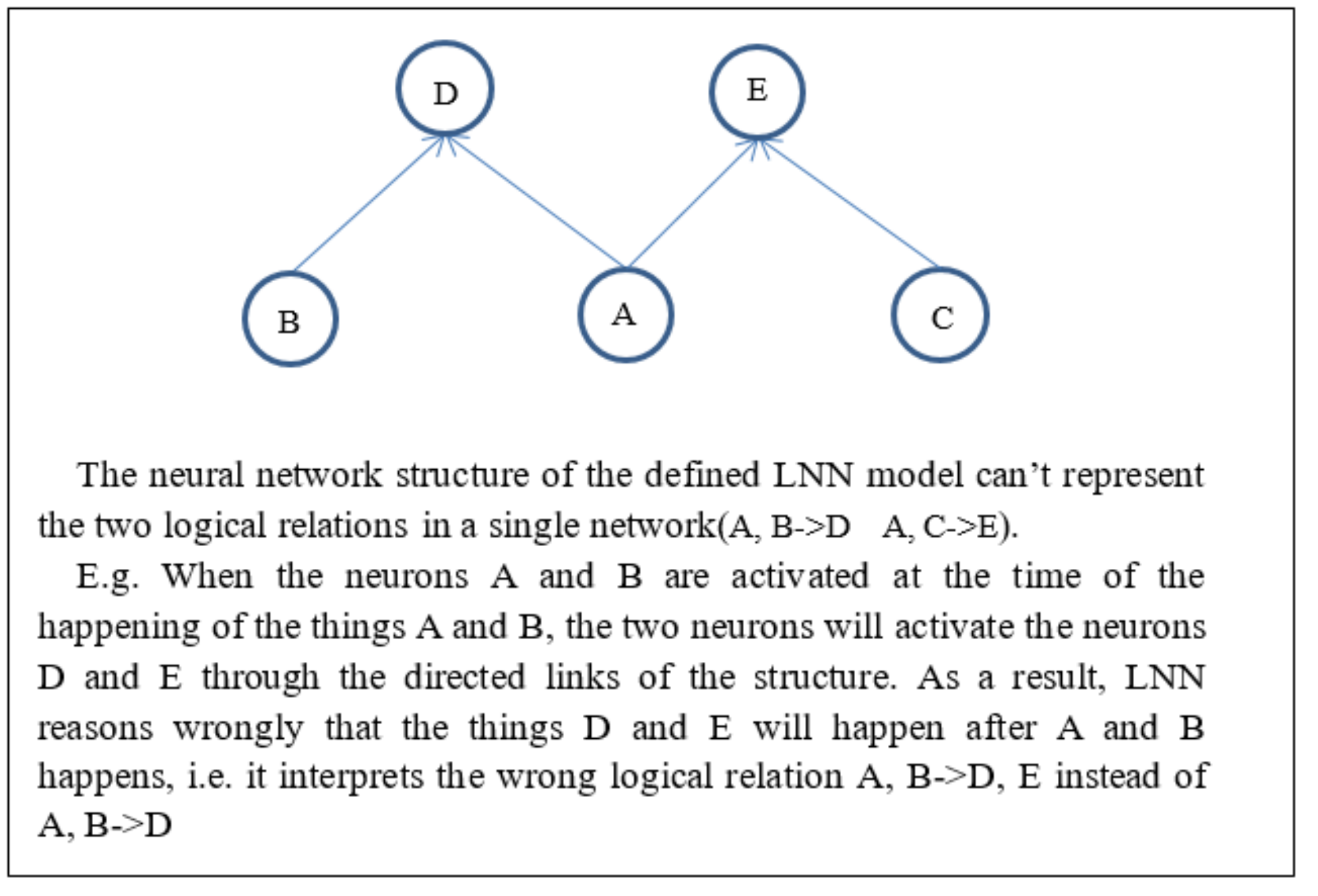}
\caption{Unable to represent multiple logical relations in a single neural network structure of the defined LNN model}
\label{LNN trouble}
\end{center}
\end{figure}

A trouble comes out that the neural network can not interpret the logical relations rightly when multiple logical relations are represented and stored in a single neural network. For example, to represent the relation {A, B$\rightarrow$ D, A, C$\rightarrow$ E} in a logical neural network, the structure is constructed as shown in Figure~\ref{LNN trouble}. However, the neural network can not interpret the logical relations again by self. As seen from the neural network structure, when the neurons A and B are activated at the time of the happening of the things A and B, the two neurons in the following will activate the neurons D and E through the directed links of the structure. As a result, LNN reasons wrongly that the things D and E will happen after A and B happens, i.e. it interprets the wrong logical relation A, B$\rightarrow$ D, E instead of A, B$\rightarrow$ D according to the current neural network structure.

\begin{figure}[!ht]
\begin{center}
\includegraphics[width=0.9\columnwidth]{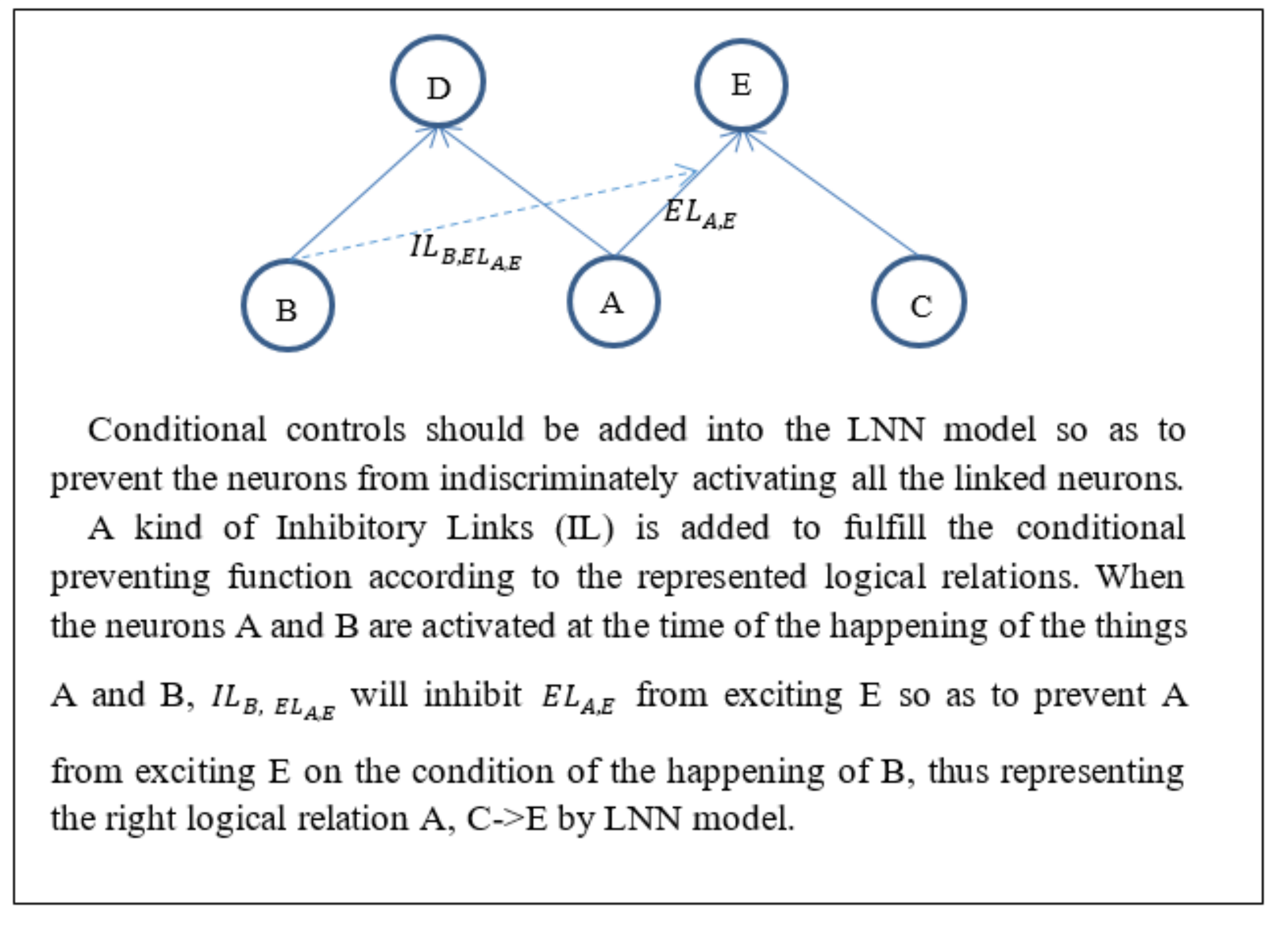}
\caption{Inhibitory links are added into the defined LNN model used as conditional prevention according to logical relations}
\label{IL}
\end{center}
\end{figure}

The reason is that the disturbance between each other appears when multiple logical relations are represented and stored in a single neural network. As shown in Figure~\ref{IL}, when the thing A and B happen, the neuron A wrongly activates the neuron E through the directed link $L_{A, E}$ formed by the logical relation A, C$\rightarrow$E. Therefore, in order to make ANN represent logical relations rightly, LNN should add conditional controls to prevent the neurons from activating the linked neurons according to the logical relations stored or memorized in the network structure. With the intention of fulfilling the effect of preventing the neurons from activating the wrong neurons through their links, this paper adds a kind of Inhibitory Links (IL) to fulfill the preventing function which is in opposition with the Excitatory Link (EL). When the neurons A and B are activated at the time of the happening of the things A and B, then A will excite D and E through its excitatory link $EL_{A, D}$ and $EL_{A, E}$. To prevent A from exciting E through $EL_{A, E}$ on the condition of the happening of B, an IL is added, and it connects the exciting link $EL_{A, E}$ to prevent A from exciting the neuron E through $EL_{A, E}$. When the neuron B is activated, the IL which connects the neuron B will inhibit $EL_{A, E}$ from exciting E, thus representing the right logical relation A, C->E by LNN model. By the interactions between neurons through expiatory and inhibitory links, this logical neural network can represent the right logical relations.

\begin{figure}[!ht]
\begin{center}
\includegraphics[width=0.9\columnwidth]{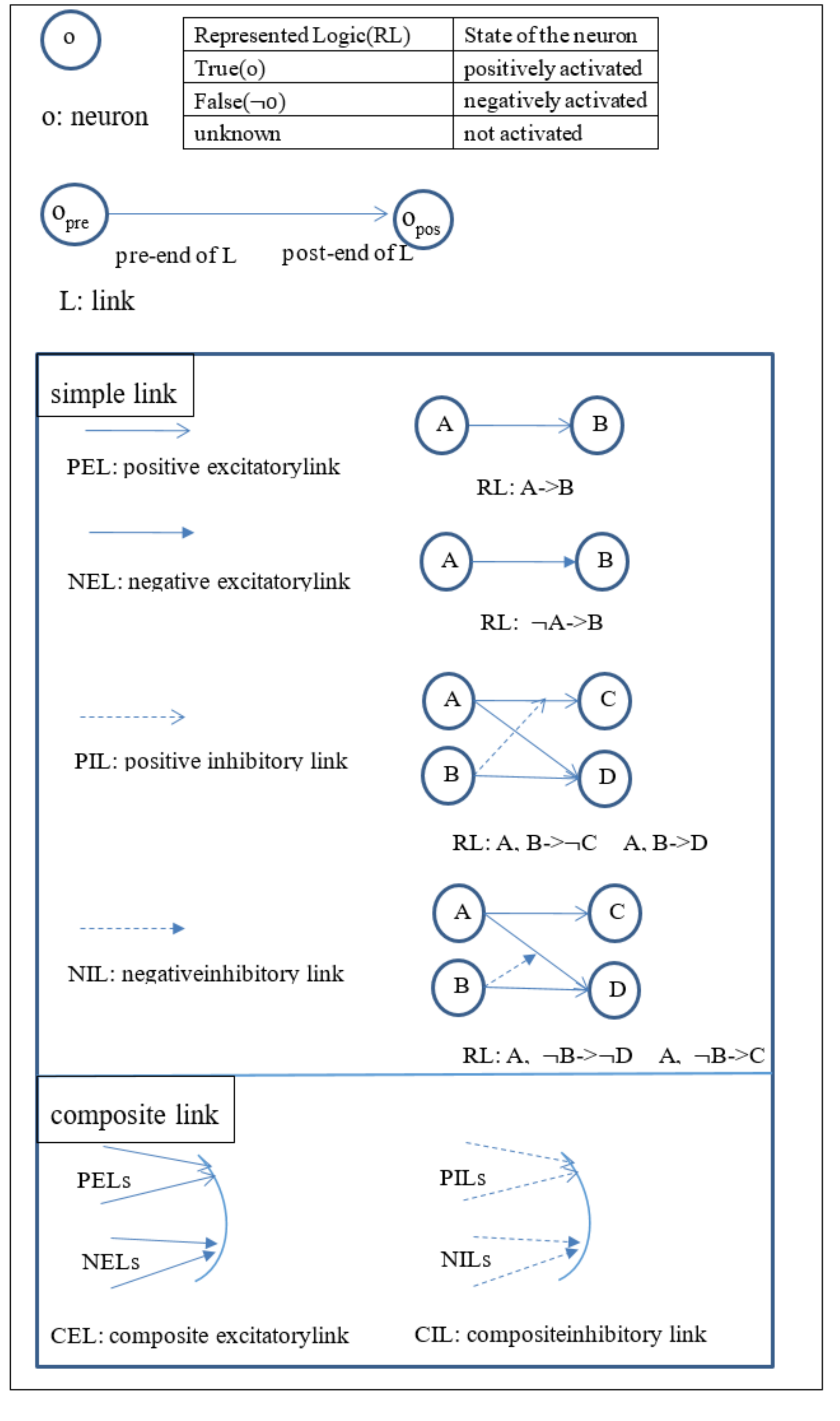}
\caption{Components of LNN to represent logical relations}
\label{components}
\end{center}
\end{figure}

Furthermore, the excitatory links from multiple pre-end neurons can be composed together and form a Composite Excitatory Link (CEL) to represent the complex logical relation. When all the pre-end neurons are activated, CEL will activate its post-end neuron. Otherwise, in the condition that partial neurons among all the pre-end neurons are activated, the CEL will not activate its post-end neuron. In the same way, the inhibitory links from multiple pre-end neurons can be composed together and form a Composite Inhibitory Link (CIL) to represent the complex logical relation. Similar with the activating way of CEL, when all the pre-end neurons are activated, CIL will inhibit its connected post-end excitatory link. Otherwise, in the condition that partial neurons among all the pre-end neurons are activated, the CIL will not make the inhibit effect. The overview of the components of the presented LNN model is shown in Figure~\ref{components}.

\begin{figure}[!ht]
\begin{center}
\includegraphics[width=0.9\columnwidth]{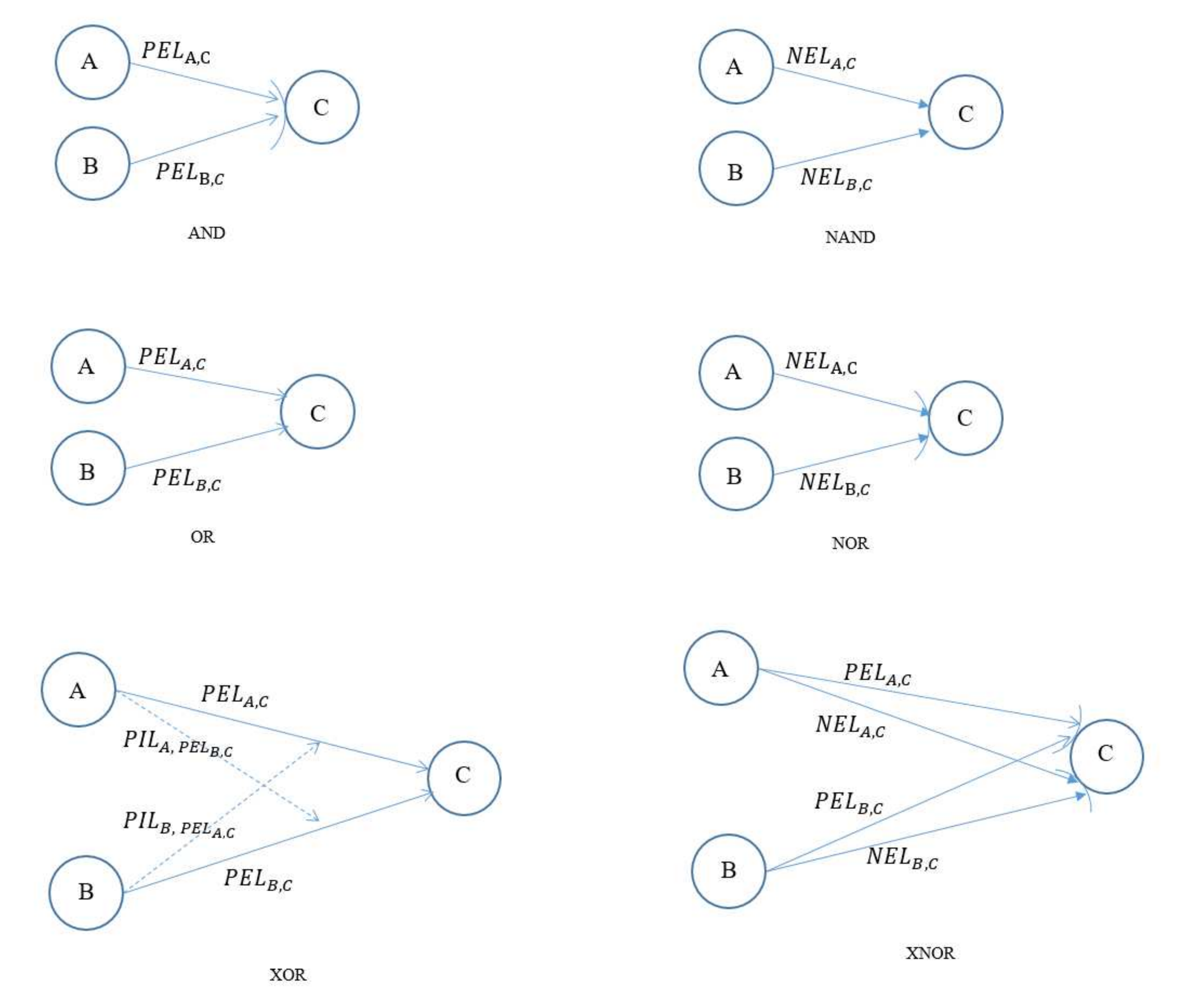}
\caption{Using only two components (neurons and links) of the LNN model to represent basic logical relations. No extra neurons and hidden layers arise.}
\label{6logic_gates}
\end{center}
\end{figure}

For the basic logical relations between two things including AND, OR, XOR, NAND, NOR and XNOR, the neural network structures using the above logical neurons and links are shown in Figure~\ref{6logic_gates}. As seeing every neural network structure in Figure~\ref{6logic_gates}, the neural network structure of this LNN model is interpretable by self. Every neuron and link has clear logical semantics, and the network structure composed by them are subsequently recognized according to the interconnection between the neurons to know what the logical relations are represented. No extra neurons and links with unknown and unclear meanings arise in the network structure of this LNN model in the process of representing logical relations.

Herein, let us take XOR as an example to see whether these structures can represent these logical relations. As shown in Figure~\ref{6logic_gates},
\begin{itemize}
  \item when the things A and B happen and perceived by the LNN model, the neurons A and B representing them are positively activated, i.e. take the values of True. Then A will excite C through its excitatory link $PEL_{A, C}$. At this time, B will inhibit $PEL_{A, C}$ to prevent A from exciting C through the inhibitory link $PIL_{B, PEL_{A, C}}$. So does A for preventing B from exciting C. Consequently, the output neuron C is not activated. The result agrees with the semantics of XOR.
  \item when only one thing like A happens and perceived by the ANN model, the neuron A representing A is positively activated, i.e. take the values of True. Then only A will excite C through its excitatory link $PEL_{A, C}$.
  \item when the things A and B don't happen, no activities are made, Consequently, the output neuron C is not activated.
\end{itemize}

Through the above explanation, it can been seen that this LNN model can represent XOR correctly.

\begin{figure}[!ht]
\begin{center}
\includegraphics[width=0.9\columnwidth]{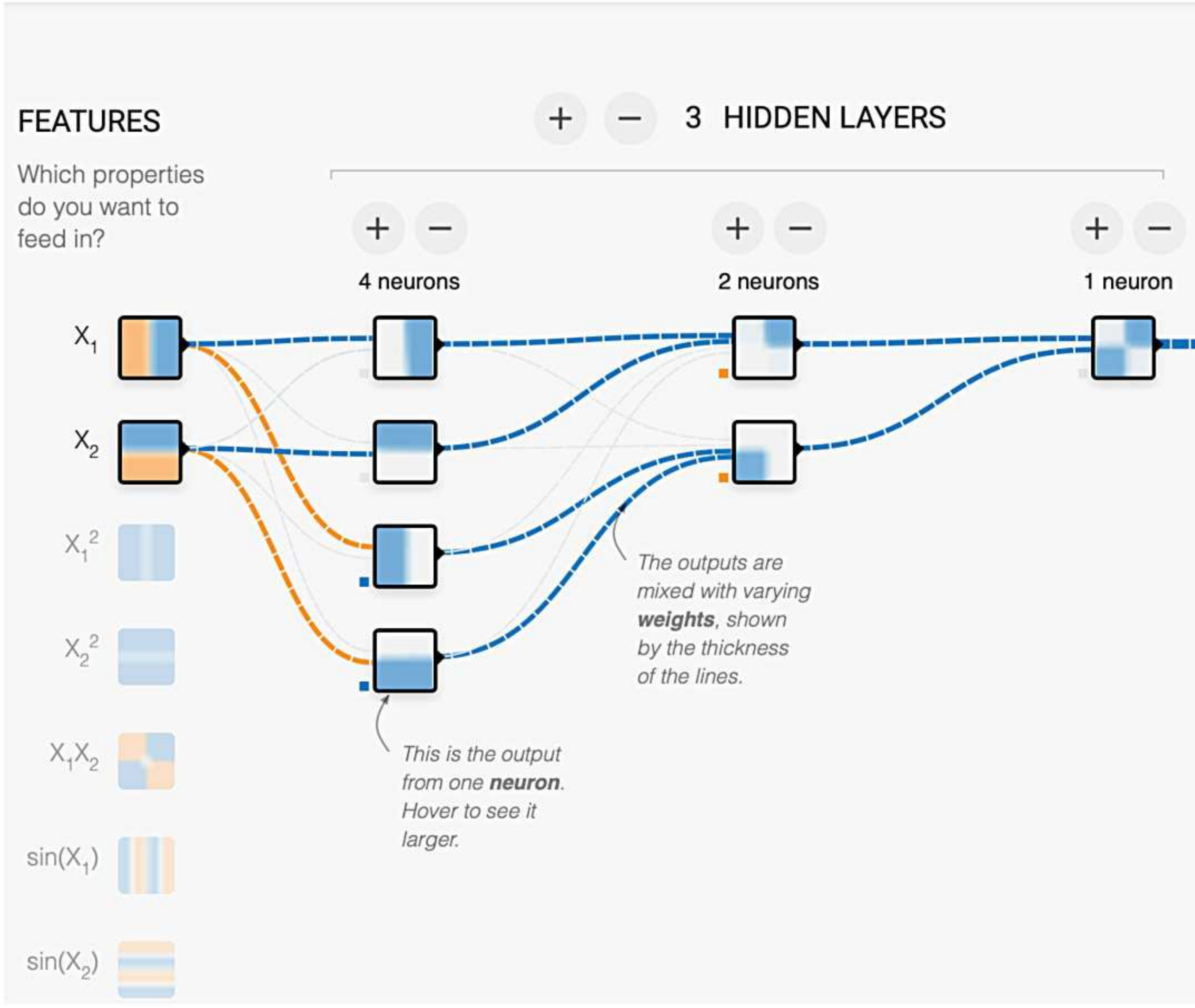}
\caption{Representing XOR by the LNN model using continuous logic. This LNN model uses 7 neurons and 3 layers to represent the basic logical relation XOR.}
\label{NLNN_XOR}
\end{center}
\end{figure}

\begin{figure}[!ht]
\begin{center}
\includegraphics[width=0.9\columnwidth]{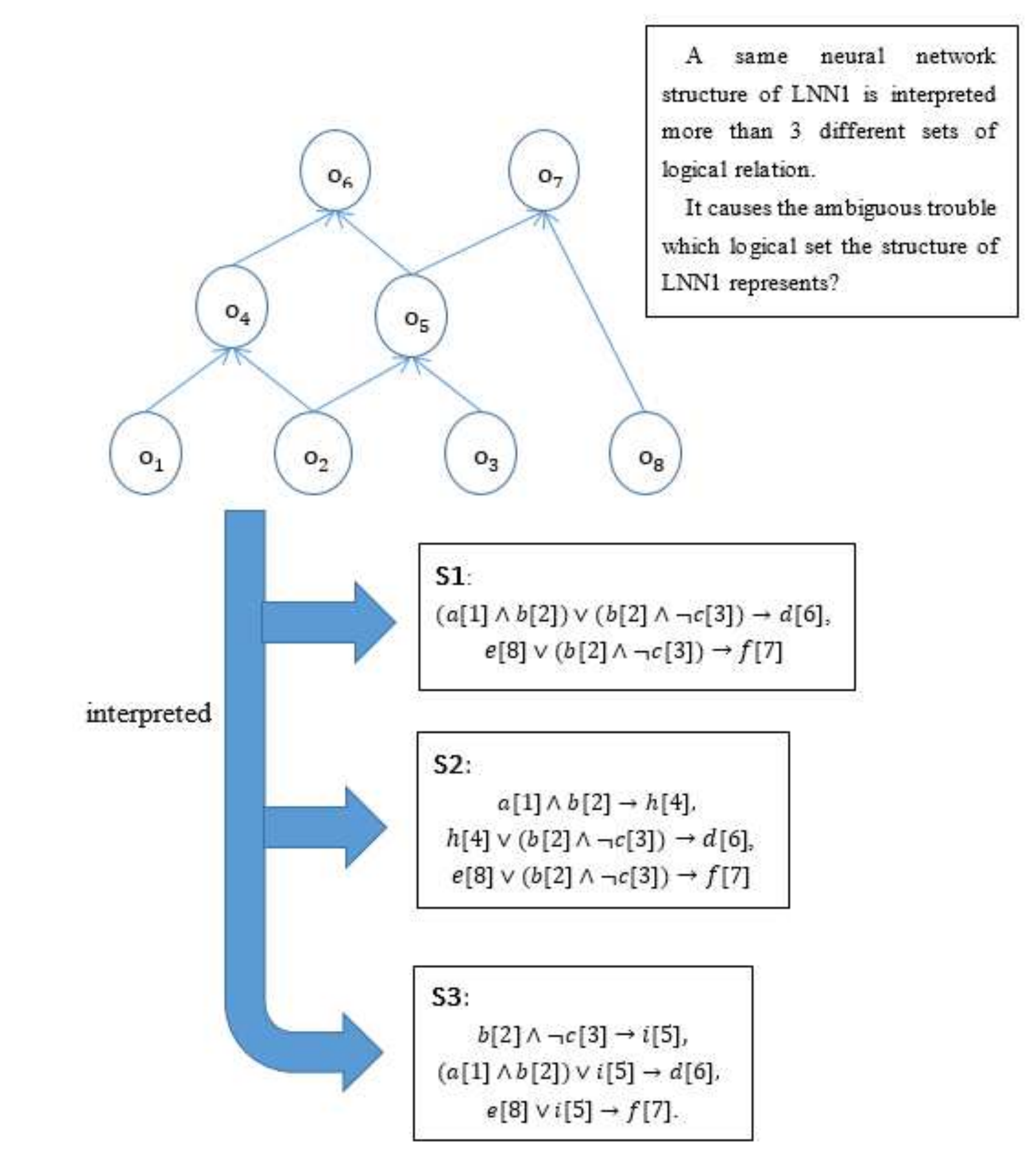}
\caption{Ambiguous interpretation of the same neural network structure of LNN1 on logical relations.}
\label{LNN troubles}
\end{center}
\end{figure}

\begin{figure}[!ht]
\begin{center}
\includegraphics[width=0.9\columnwidth]{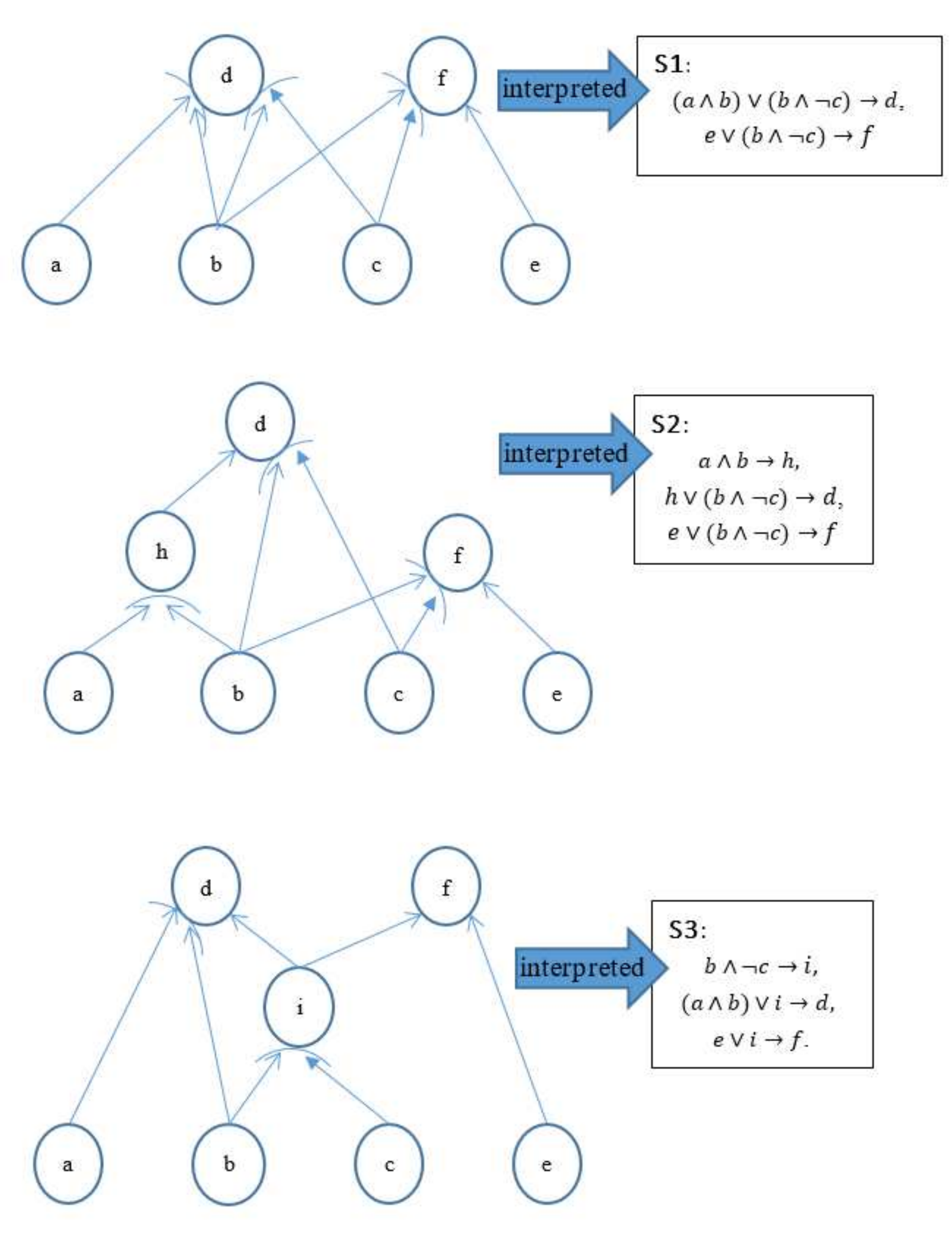}
\caption{3 different sets of logical relation are interpreted by 3 different neural network structures of the presented LNN model, keeping the same layer structure information with the logical relations it represents}
\label{LNN solution}
\end{center}
\end{figure}

By comparison with the recent work on the LNN model \cite{Csiszar} indicated by LNN1, LNN1 needs extra neurons added in the hidden layers so as to represent the logical relations such as XOR as shown in Figure~\ref{NLNN_XOR}. LNN1 defines logical neurons by setting weights and biases as shown in Table~\ref{Weights and biases}. In order to represent the basic logical relation XOR, LNN1 uses 7 neurons and 3 layers at a price. Moreover, when using LNN1 to represent the set of multiple logical relations, it may make ambiguousness to interpret the neural network structure because there will be the same neural network structure for representing the different logical relations. For example, for the set of logical relations S1 \{$(a \bigwedge b) \bigvee (b \bigwedge \neg c) \rightarrow d, e \bigvee (b \bigwedge \neg c) \rightarrow f$\}, the logical neural network structure is constructed in Figure~\ref{LNN troubles}. In Figure~\ref{LNN troubles}, the number in the square bracket indicates which neuron representing the thing. E.g. a[1] means the neuron o1 represents the thing a. In order to represent the set of logical relations S1, LNN1 uses the neuron $o_{1}$ to represent the thing a appearing in the logical relation $(a \bigwedge b) \bigvee (b \bigwedge \neg c) \rightarrow d$, and LNN1 uses $o_{2}$ to represent b. Next LNN1 uses $o_{3}$ to represent $\neg c$. In order to represent the logical relation $\bigwedge$, LNN1 uses a new neuron $o_{4}$. So does $o_{5}$. Further, LNN1 uses $o_{6}$ to represent the logical relation $\bigvee$ and the the thing d at the same time. So does $o_{7}$. Finally, LNN1 uses $o_{8}$ to represent e. By now, the neural network structure using LNN1 model to represent S1 is constructed as shown in Figure~\ref{LNN troubles}. However, for another logical relation S2 \{$a \bigwedge b \rightarrow h, h \bigvee (b \bigwedge \neg c) \rightarrow d, e \bigvee (b \bigwedge \neg c) \rightarrow f$\}, the network structure is also the same structure in Figure~\ref{LNN troubles}. In addition, for the logical relation S3 \{$b \bigwedge \neg c \rightarrow i, (a \bigwedge b) \bigvee i \rightarrow d, e \bigvee i\rightarrow f$\}, the network structure is also the same structure in Figure~\ref{LNN troubles}. There are still other sets of logical relations represented by this same neural network structure using LNN1 model, which are not mentioned anymore for the concise statement. The ambiguousness appears when interpreting the neural network structure of LNN1. The reason is the design style of putting multiple responsibilities on neurons. In more words, in this design style, a neuron holds the multiple responsibilities including representing the things and representing the logical relations as well. Sometimes the neuron only represents a thing, sometimes the neuron only represents a logical relation, and sometimes the neuron represents both a thing and a logical relation. To avoid putting multiple responsibilities on the neuron, in this paper, the neuron is designed only for representing the thing while the representation of logical relations between the things are wholly addressed by the links. It is natural and more intuitive of this design style for the reason that: when there are the logical relations between the things, new links rather than new neurons should appear to represent the logical relations the things by the way of connecting the neurons representing these things. Moveover, in this design style, the neural network structure keeps the same layer structure information with the logical relations it represents, which can be seen in Figure~\ref{LNN solution}. The logical neural network model in the paper has no ambiguousness to interpret the neural network structure aforementioned.
\begin{table}[!ht]
\caption{Weights and biases for representing XOR}
\label{Weights and biases}
\centering
\begin{tabular}{|c||c|c|c|}
  \hline
  RL & $w_{1}$ & $w_{2}$ &b\\\hline
  x AND y & 1 & 1 & -1\\\hline
  x OR y & 1 & 1 & 0\\\hline
  NOT(x) & -1 & & 1\\\hline
  x AND y & 1 & 1 & -1\\\hline
  NOT(x) AND NOT(y) & -1 & -1 & 1 \\
  \hline
\end{tabular}
\end{table}
\section{Neural network structure of logical relations constructed based on these novel logical neurons and links}
\label{structure}
In this section, we will take the same set of logical relations used in constructing the neurule base \cite{Ioannis} to construct the LNN based on these novel logical neurons and links. Through the demonstration of the the network structure of the LNN model constructed and by comparison with the neurule model, it can be seen that logical relations can be interpreted more easily and intuitively from the neural network structure of the presented LNN model. The set of logical relations used in \cite{Ioannis} is about the medical diagnosis of bone diseases, which is shown by the neurule model in Table~\ref{neurules}. The neurules R1 - R3 divides patients into three classes of the patients according to the gender and age. The neurules R4 - R7 make preliminary bone diagnoses based on four symptoms ('pain', 'antinflam-reaction', 'fever' and 'joints-pain') and the 'patient-class'. The diagnostic results includes two types of the diseases('arthritis' and 'primary-malignant').
\begin{table*}[!ht]
\caption{Logical relations about the medical diagnosis of bone diseases represented by the neurule model}
\label{neurules}
\centering
\includegraphics[width=0.9\textwidth]{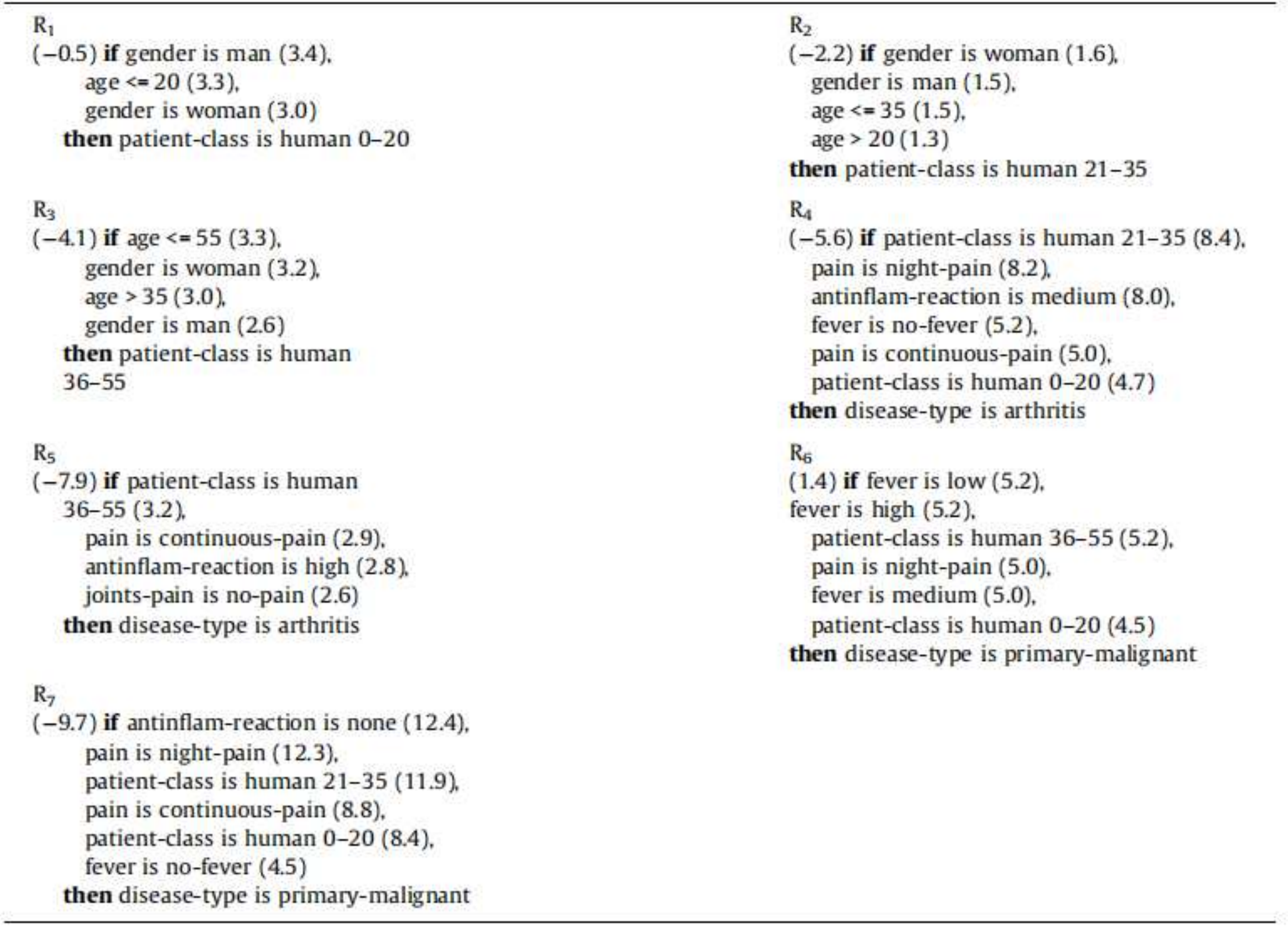}
\end{table*}

\begin{figure}[!ht]
\begin{center}
\includegraphics[width=0.9\columnwidth]{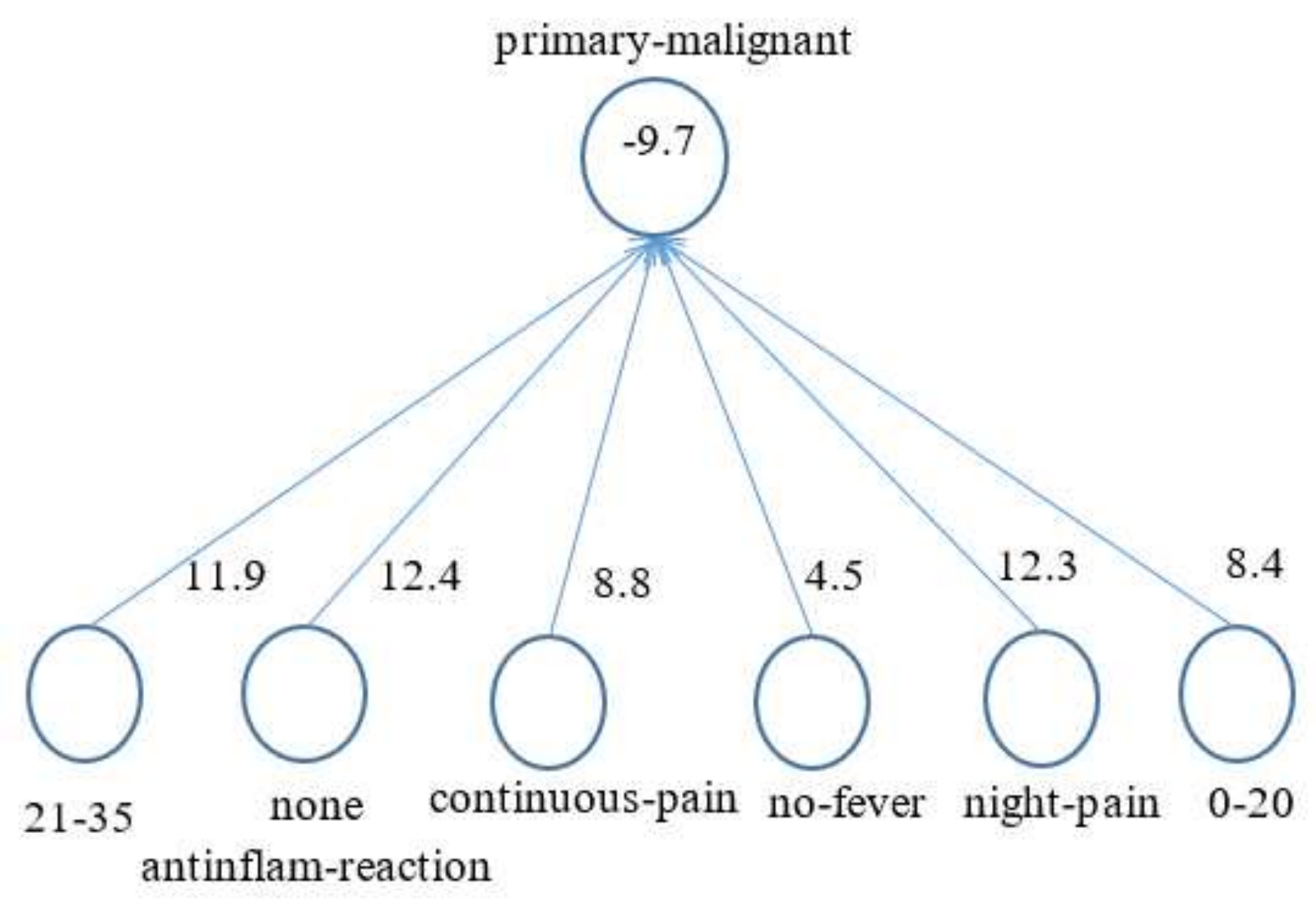}
\caption{Neurule R7 to represent to the above 4 symbolic rules}
\label{neuruleR7}
\end{center}
\end{figure}

Herein, to put a question to the readers, what logical relations can you read out from this neural network structure of the neurules such R2 in Figure~\ref{neurule} and Table~\ref{neurules} or the more complex neurule R7 in Figure~\ref{neuruleR7} and Table~\ref{neurules}? Can logical relations be interpreted directly from this neural network structure of the neurule model? Actually, a neurule doesn't map directly to one logical relation, which makes the network structure of the neurule model uneasy to interpret. For example, the neurule R2 corresponds to the following two symbolic rules:
\begin{enumerate}
\item if gender is woman,

  and 20 $<$age $\leq$ 35

then patient-class is human 21-35
\item if gender is man,

  and 20 $<$age $\leq$ 35

then patient-class is human 21-35
\end{enumerate}

the neurule R7 corresponds to the following 4 symbolic rules:
\begin{enumerate}
\item if antinflam-reaction is none,

  patient-class is human 21-35,

  pain is continuous-pain,

  and fever is no-fever

  then disease-type is primary-malignant
\item if antinflam-reaction is none,

  patient-class is human 21-35,

  and pain is night-pain,

  then disease-type is primary-malignant
\item if antinflam-reaction is none,

  patient-class is human 0-20,

  pain is continuous-pain,

  and fever is no-fever

  then disease-type is primary-malignant
\item if antinflam-reaction is none,

  patient-class is human 0-20,

  pain is night-pain,

  and fever is no-fever

  then disease-type is primary-malignant
\end{enumerate}

\begin{figure}[!ht]
\begin{center}
\includegraphics[width=0.9\columnwidth]{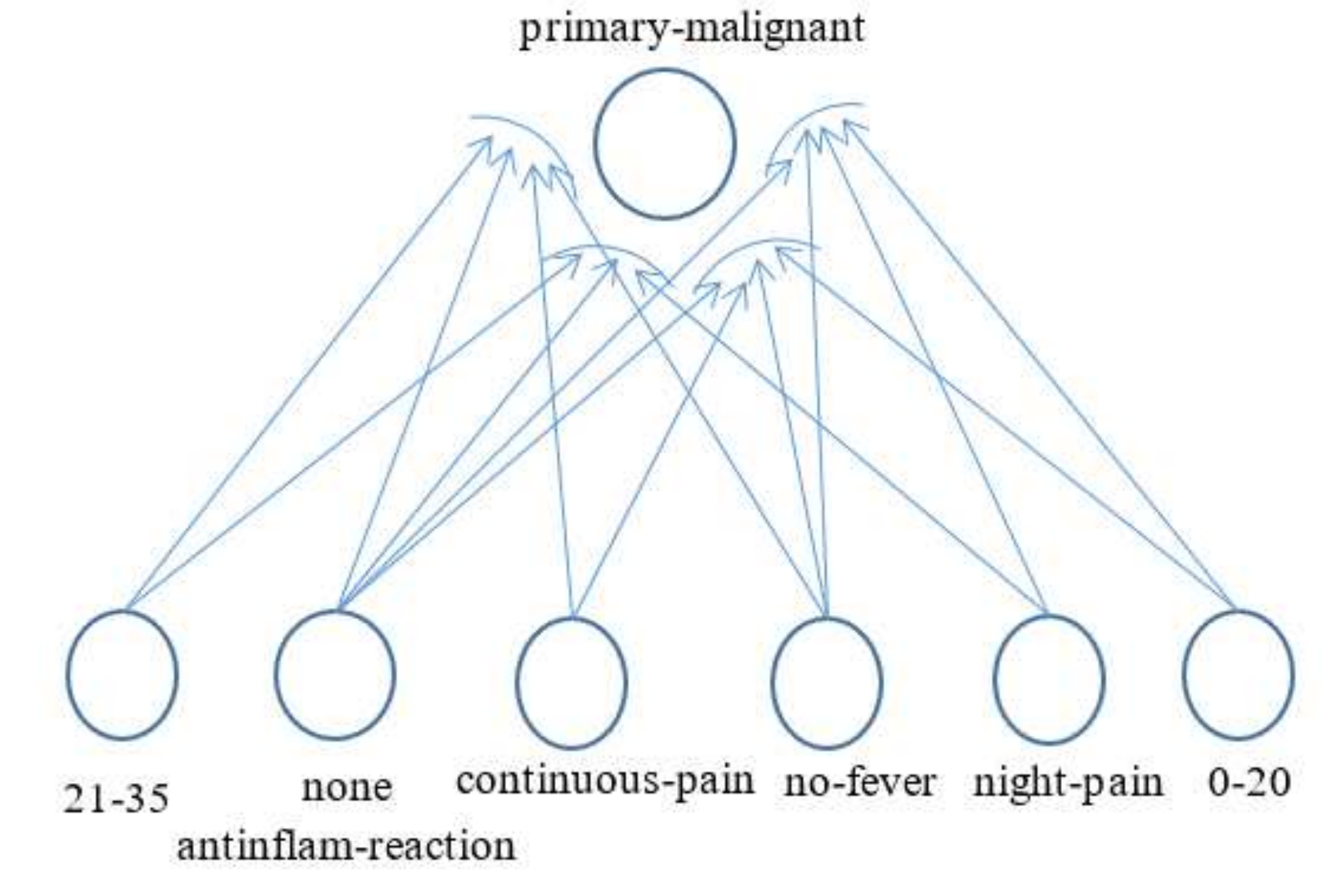}
\caption{The presented LNN model to represent to the above 4 symbolic rules}
\label{LNNR7}
\end{center}
\end{figure}

By comparison with the neurule model, the LNN model presented in this paper maps these logical relations directly to the corresponding neurons and links shown in Figure~\ref{LNNR7}. The neural network structure of the presented LNN model is constructed explicitly according to the 4 symbolic rules by using 4 composite excitatory links. Different from the neurule model, this model keeps the information of these logical relations, thus its neural network structure is interpretable and the logical relations can be directly read out from the neural network structure of this model. Readers can even intuitively and easily read out the 4 symbolic rules from the neural network structure of the presented LNN model in Figure~\ref{LNNR7}. The LNN model doesn't reduce the links which is different from the neurule model, and this way will keep the information of the represented logical relations. It is useful when adjusting the network model when the logical relations changes.

For example, assuming the first logical relation in the above 4 symbolic rules indicated by R7.1 is removed from the set for some reason such as discovering R7.1 is not right in the diagonal process, therefore, in order to represent new logical relations, the neurule model has to adjust the bias or weights. The sum S of R7.1 is S(R7.1)=-9.7+12.4*1+12.3*(-1)+11.9*1+8.8*1+8.4*(-1)+4.5*1=7.2 according to the bias or weights assigned in the neurule R7 in Table~\ref{neurules}.
In order to make this neurule not to represent R7.1, one way is to increase the bias from 9.7 to make S(R7.1) $<$0 like setting the bias 17(9.7+7.3), but it will disturb the neurule R7 to represent the other logical relation like R7.4. The sum of R7.4 in the condition of unchanged bias is S(R7.4)=-9.7+12.4*1+12.3*1+11.9*(-1)+8.8*(-1)+8.4*1+4.5*1=7.2. If the bias is set to 17 in order not to represent the logical relation R7.1, the neurule R7 afterwards will be unable to represent R7.4 and R7.3 as well. Another adjusting way is to adjust the weights. Here use a case of increasing the weight of "night-pain" for illustration. If the weight of "night-pain" is set 19.6(12.3+7.3), the neurule R7 will be unable to represent R7.3. There are still other ways to adjust this neurule model by changing the bias and weights simultaneously. However, it can been seen that the adjusting process is uneasy.

By comparison, due to its direct mapping from logical relations to its neural network structure, the LNN model presented in this paper just removes the corresponding links, and then it realizes the new logical representation of the changed logical relations. The reason causing the uneasy adjustment for the neurule model is high coupling, the term which we borrow from software architecture design or soft engineering. Multiple logical relations tightly compose together to form a neurule. If one or more logical relations changes, it may disturb the right representation of other logical relations in the same neurule after this neurule adjusts its bias or weights. The neurule model reduces the links at the cost of losing representing information of logical relations. When the network structure of the neurule model adjusts for the changed logical relations, it will disturb the right representation of other logical relations.
\section{Experiment and demonstration}
\label{Experiment}
In this section, in order to test whether the LNN model presented in this paper can represent logical relations, the experiment will be carried out further with the datasets. The experiment process is the following: the LNN model constructs its network structure according to every record in a dataset. The feature attributes of each record are used as the condition of an \emph{if-then} rule, the conclusion in each record is used as the conclusion of the \emph{if-then} rule. The LNN model will create corresponding neurons to represent the things in the attributes and conclusions, then creates the directed neural links from the neurons representing the things in the attributes to the neurons representing the things in the conclusions. The LNN model represents and stores the logical relations in the datasets into its network by this constructing way.

In the experiment, after the new neural network structure is constructed and adjusted continuously and increasingly according to another new record in the dataset, the LNN model will be required to read out again all the previous logical relations it stored so as to examine whether the logical relations it represents and stores are the same with those logical relations in the dataset. In more words, the LNN model is constructed and adjusted continuously and increasingly along with each record in the dataset one by one. The constructing and adjusting algorithm of the neural network structure is implemented in Java. If the LNN model gives the conclusions rightly for the values of the attributes in the previous records, then it has represented and stored these logical relations rightly. Next it will represent and store another new record in the dataset by constructing and adjusting its neural network structure. Otherwise, the LNN model repeats this process until it represents and stores all the logical relations in the dataset or stops in the limited steps if it cannot gives the right conclusions all. If the LNN model can read out again all the logical relations in the dataset in the limited steps, then it will be examined that the LNN model can represent and store logical relations in the dataset.

Two datasets are used. The fist dataset is called Mushroom \cite{UCI}, which is taken from UCI Machine Learning Repository. The dataset describes the feature attributes of various kinds of mushrooms and gives the conclusions whether the mushrooms are edible or not. It is an useful function to memorize the feature attributes of the mushrooms and their corresponding edibility for outdoors men. In the experiment, the dataset is modified for incremental tests. The smallest sub-dataset consists of 10 feature attributes with 25 records. The characteristic of this sub-dataset is that: on the one hand, the number of the combination of feature attributes and conclusions is enormous after multi multiplications among 10 feature attributes and the conclusions, and they form a great space of logical relations; on the other hand, the various records in the sub-dataset is like individually separate particles scattered in this great space. Therefore, it is practical and useful to memorizes these specific records of logical relations rightly among enormous numbers of the combination, especially in a critical situation like picking mushrooms outdoors or medical diagnose.

In the experiment, the paper will examine this LNN model whether it can memorize these specific records of logical relations,i.e, whether it can give the conclusions rightly given the values of the attributes in the records. Next, this paper increases the feature attributes to 15, 20 and increases the number of records to 50, 75 to examine whether the LNN model is still workable along with the size increment of the dataset. After performing the experiments, Table~\ref{Results} presents the experimental result shows the model represents and stores all the logical relations in the dataset, i.e. that it can give the conclusions rightly in the conditions of the values of the attributes in the records.
\begin{table}[!ht]
\caption{Experimental results of the LNN model for logical representation}
\label{Results}
\begin{center}
\begin{tabular}{|c||c||c|}
\hline
Num(attributes) & Num(mushrooms) & recallings\\\hline
10 & 25 & all \\\hline
15 & 50 & all \\\hline
20 & 75 & all \\\hline
\end{tabular}
\end{center}
\end{table}

Furthermore, in order to show the practicability and university of this logical neural network model, another different dateset is used from the medical domain called SPECT Heart\cite{SPECT+Heart} from Machine Learning Repository as well. The dataset describes the feature attributes of SPECT of the hearts and gives the conclusions whether the hearts are normal or abnormal. The records with indecisive conclusions are removed from the dataset. Through performing the experiment similarly, the experimental result shows the LNN model is still workable, which means it represents and stores all the logical relations in the dataset, i.e. that it can give the conclusions rightly in the conditions of the values of the attributes in the records.

\begin{figure}[!ht]
\begin{center}
\includegraphics[width=0.9\columnwidth]{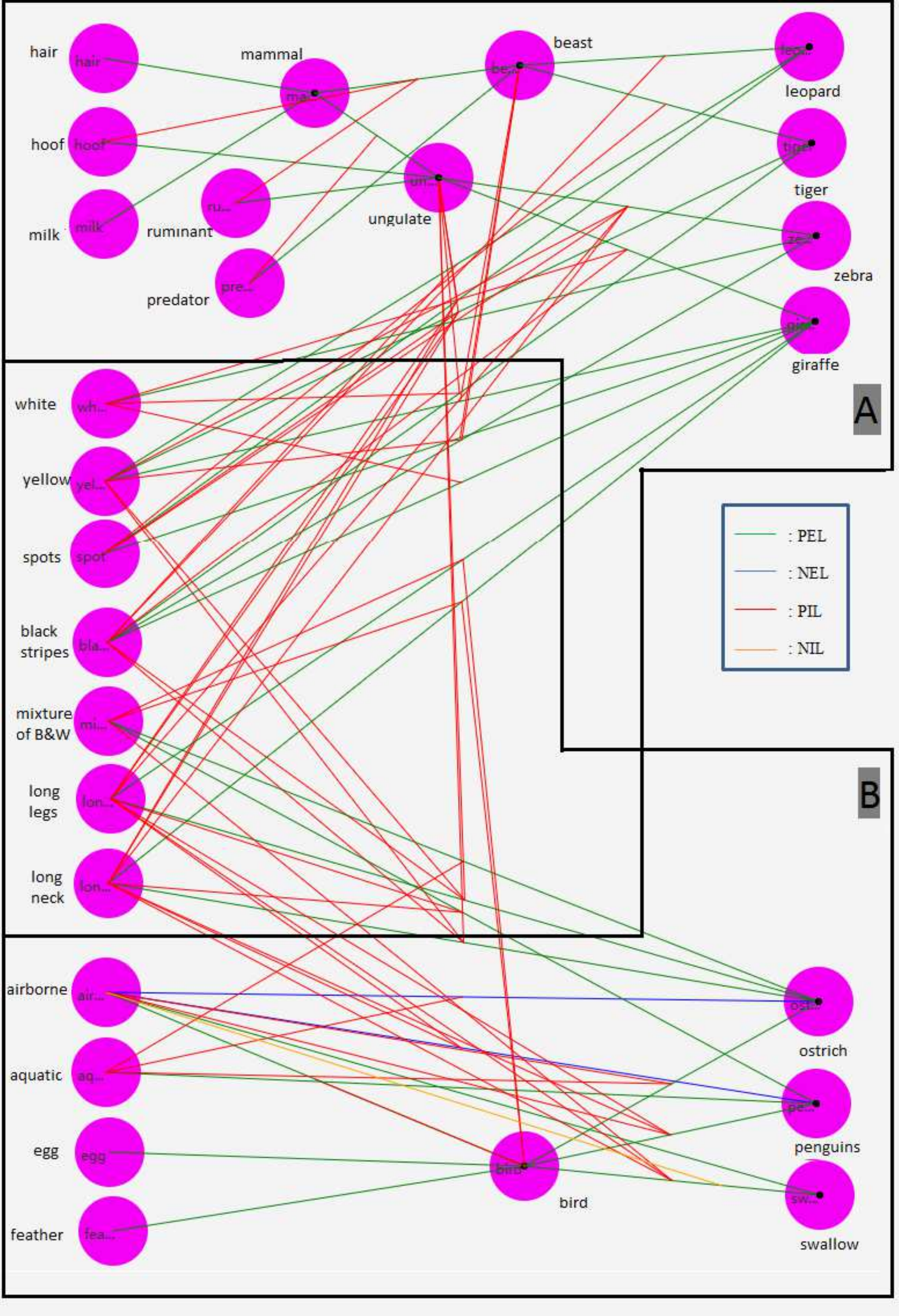}
\caption{Integrated Neural network structure to represent logical relations of animals including mammals and birds}
\label{animal_zone}
\end{center}
\end{figure}

An demonstration of using the LNN model is presented to the advantages of this LNN model in logical representation. The LNN model represents and stores logical relations about the domain of animals in appendix~\ref{Supplements}. The constructed neural network structure is graphically depicted in Figure~\ref{animal_zone}. As can be seen from Figure~\ref{animal_zone}, this LNN model produces more natural mapping from logical relations to neural network structure. In addition, the neural network structure of the LNN model also represents and stores the hierarchy in the logical relations. There is no neurons and links which have no ambiguously meanings to users. Every neurons and links are created according to logical relations. Neurons represent things while links represents the logical relations between the things. The logical relations can be read out from the network structure.

For example, according to the demonstrative rule base in appendix~\ref{Supplements}, the neurons called "hair" and "mammal" are created to represent the things "hair" and "mammal" appearing in the logical relation R1, and a link connects them from the neuron "hair" to "mammal" so as to represent R1. Similarly, the neurons called "predator" and "beast" are created to represent the things "predator" and "beast" appearing in R3, and two links connect them from the neurons "mammal" and "predator" to  "beast" so as to represent R3. The LNN model constructs its network structure according to the logical relations, and it can be seen intuitively that the neural network structure also has the same hierarchies with those of the logical relations. When the LNN model perceives a condition ("mammal", "predator"), the neuron M representing "mammal" will excite the neurons B and U ("beast" and "ungulate") through M's PELs. The neuron P representing "predator" will excite the neurons B. At the same time, the neuron P's PIL inhibits M's PEL from exciting the neuron U. Through these multiple exciting and inhibiting interactions among neurons through links, only the neuron B is activated, then this LNN model finally gives the conclusion that the animal is a beast under the condition ("mammal", "predator"). It is similar for the other logical relations, it can be manually verified if interested. By illustrating the interactions of between neurons by links, it can be seen the LNN model in this paper has represented the logical relations rightly.

On the contrary, the hidden neurons introduced in numerical NN models have no exact meanings to users, which will downgrade the clearness and naturalness of the neural network model for use in representing logical relations. The neural network structure of the numerical NN models will confuse users and raise a problem of comprehensibility since these hidden neurons can't correspond to conditions and conclusions of logical relations.
\section{Conclusion}
\label{Conclusion}
In this paper, a LNN model is presented which maps more directly from logical relations into neural network structure. The LNN model constructs its network structure by creating neurons and links on demand according to logical relations. An attractive feature of the model is that compared to other similar LNN models like the two aforementioned models, no more extra neurons and links having inexact meanings exist in the network structure. The network structure of this neural network is dynamically constructed following the perceived logical relations other than the predefined fixed network structure. This dynamical characteristic makes the model dynamically adjust its network structure along with the change of logical relations in the rule base. Logical relations can be read out from the neural network structure more naturally and easily due to this direct mapping characteristic, which is shown and explained in details through the demonstration of using this LNN model to represent logical relations and the neural network structure graphically depicted in Figure~\ref{animal_zone}. This work presents a novel way of how to represent and store logical relations in the format of neural network with the more direct mapping way. We hope this work can contribute some ideas for colleagues in this research topic.


%

\section*{Acknowledgment}
This paper was supported by NSFC (National Natural Science Foundation of China) Grant no. 61702356.

\ifCLASSOPTIONcaptionsoff
  \newpage
\fi



%

\appendices
\section{}
\label{Supplements}
In the following, it is a simple rule library as example which contains 14 logical relations. The LNN model presented in Figure~\ref{animal_zone} memorizes them and forms a knowledge graph through the interconnection structure of the neural network.
\begin{enumerate}
  \item If an animal has hair,  then it is mammal
  \item If an animal produces milk,  then it is mammal
  \item If a mammal is predator,  then it is beast
  \item If a mammal has hoof,  then it is ungulate
  \item If a mammal is ruminant, then it is ungulate
  \item If an animal has feather, produces egg,  then it is bird
  \item If an animal airborne, then it is bird
  \item If a beast is yellow and spots,  then it is leopard
  \item If a beast is yellow and black strips,  then it is tiger
  \item If an ungulate has long neck, long leg, yellow and spots, then it is giraffe
  \item If an ungulate is white and black strips,  then it is zebra
  \item If a bird cannot airborne, has long neck, long legs, and is mixture of black and white,  then it is ostrich
  \item If a bird cannot airborne, can aquatic, and is mixture of black and white, then it is penguin
  \item If a bird can airborne,  then it is swallow
\end{enumerate}

These relations often appeared as example and appeared in AI-related papers and books such as Neural Networks and Learning Machines written by Haykin \cite{Simon}.



\end{document}